%% file: SMNARX_v2.tex
\documentclass[conference]{IEEEtran}
\IEEEoverridecommandlockouts
\usepackage{cite}
\usepackage{mathtools}
\usepackage{amsmath,amssymb,amsfonts}
\usepackage{algorithmic}
\usepackage{graphicx}
\usepackage{textcomp}
\usepackage{algorithmic}
\usepackage{algorithm}
\usepackage{url}
\usepackage{color}
\def\BibTeX{{\rm B\kern-.05em{\sc i\kern-.025em b}\kern-.08em
    T\kern-.1667em\lower.7ex\hbox{E}\kern-.125emX}}
\begin{document}

\title{Estimation of Switched Markov\\ Polynomial NARX models}
	

\author{\IEEEauthorblockN{Alessandro Brusaferri\textsuperscript{a,b}, Matteo Matteucci\textsuperscript{b}, Stefano Spinelli\textsuperscript{a,b}}
	\IEEEauthorblockA{\textit{\textsuperscript{a}CNR-Institute of Intelligent Industrial Technologies and Systems for Advanced Manufacturing, Milan, Italy} \\
		\textit{\textsuperscript{b}Politecnico di Milano - Department of Electronics, Informatics and Bioengineering, Milan, Italy}\\
		$name.surname$@stiima.cnr.it, $name.surname$@polimi.it}
	
}


\maketitle
\begin{abstract}
This work targets the identification of a  class of models for hybrid dynamical systems characterized by nonlinear autoregressive exogenous (NARX) components, with finite-dimensional polynomial expansions, and by a Markovian switching mechanism. The estimation of the model parameters is performed under a probabilistic framework via Expectation Maximization, including submodel coefficients, hidden state values and transition probabilities. Discrete mode classification and NARX regression tasks are disentangled within the iterations. Soft-labels are assigned to latent states on the trajectories by averaging over the state posteriors and updated using the parametrization obtained from the previous maximization phase. Then, NARXs parameters are repeatedly fitted by solving weighted regression subproblems through a cyclical coordinate descent approach with coordinate-wise minimization. 
Moreover, we investigate a two stage selection scheme, based on a $\ell1$-norm bridge estimation followed by hard-thresholding, to achieve parsimonious models through selection of the polynomial expansion. The proposed approach is demonstrated on a SMNARX problem composed by three nonlinear sub-models with specific regressors.
\end{abstract}

\input{introduction}
\input{problemFormulation}

\input{Method}

\input{Results}

\input{Conclusion}

\bibliographystyle{ieeetr}
\bibliography{BLSTM}

\end{document}

%% file: introduction.tex
\section{Introduction}
Hybrid systems provide an integrated framework to model complex interactions between discrete and continuous behaviors, as it is the case for Cyber Physical Systems \cite{Yuan2019}. 
Piecewise and Switched model forms are widely considered for identification purpose. While the former characterizes the discrete modes by partitioning the input/state space, the latter considers arbitrarily switching patterns, both in terms of the transition instants and discrete state values \cite{lauer_book}. 
 
A substantial amount of research works has been dedicated over the past decades to the special subclass of affine dynamics. A non-exhaustive list of the most popular approaches 
includes sparse regression methods, sum-of-norms regularization, recursive methods, clustering based techniques, the bounded error approach and the mathematical programming methods (see e.g,\cite{lauer_book} and references therein).

Despite being crucial to address complex applications, far fewer studies have been dedicated to the general case of Switched nonlinear dynamical systems \cite{BIANCHI2020100818}. A first attempt has been performed in \cite{10.1007/978-3-540-78929-1_24} where kernel regression and Support Vector Machines are employed, but resulting feasible just for rather limited dataset sizes. Hence, several extensions have been investigated, including differentiable approximations using the products of error estimator, Least-Squares Support Vector Machines, Feature Vector Selection, Kernel Principal Component Regression and Reduced KPCR by Incomplete Cholesky Decomposition \cite{6060916}. Authors in \cite{5717199} proposed an extension of the error sparsification approach to the nonlinear case. Then, a convex relaxation of the $l0$-norm is employed, providing optimality guarantees for noiseless data. Robust sparsity is introduced in \cite{10.1145/2461328.2461336} to address common noisy conditions, followed by the iterative solution of Support Vector Regression subproblems. 

While aforementioned studies 
exploit nonparametric techniques (i.e, kernels), a parametric perspective is proposed in \cite{BIANCHI2020100818} by adopting finite-dimensional parameterized polynomial expansions, enabling the characterization of a broad range of nonlinear systems using a small number of parameters while improving model interpretability.
Such representations suffer curse of dimensionality, therefore the integration of model structure selection (MSS) facilities is required for their application in practice. An iterative method is developed to alleviate the combinatorial complexity of the resulting mixed integer optimization problem, where mode  assignment and MSS are tuned through a randomized strategy, followed by refinement of the switching instants.  

Despite the specific class of model adopted to represent the continuous dynamics, a common feature of the reported approaches resides in the treatment of the switching mechanism. Indeed, the observations are typically segmented - by inferring the hidden active states - to identify the submodels, but the stochastic information behind the transitions is neither reconstruct nor exploited \cite{PIGA2020109126}. In various application fields, as e.g, in the operation of industrial processes, mode transitions follows certain patterns that can be exploited to improve predictions \cite{JIN2012436}. To this end, the discrete state evolution can be modeled by Markov chains, leading to Jump Markov Systems \cite{GARULLI2012344}.  
As for the piecewise and switching models above, most studies in this context have focused the linear subclasses, i.e, Jump Markov Linear Systems, jump BoxJenkins and Switched Markov AutoRegressive eXogenous systems (see e.g, \cite{PIGA2020109126},\cite{JIN2012436},\cite{GARULLI2012344}). Jump Markov Nonlinear Systems are investigated in \cite{6994863}, proposing a  recursive maximum likelihood approach based on a Rao-Blackwellized particle filter.

In this work, we investigate a nonlinear extension of the Switched Markov Autoregressive eXogenous systems. Specifically, we consider a parametric approach to represent the continuous dynamics by including polynomial expansions, thus obtaining a Switched Markov Nonlinear ARX system (labeled SMNARX hereafter for notation simplicity). The estimation of the overall model parameters, including NARXs, hidden state values and transition probabilities, is performed under a probabilistic framework via Expectation Maximization. Then, mode-wise regressors are sparsified within the disentangled weighted regression subproblems, where the soft-assignments to the latent states are iteratively updated through the posterior estimations until convergence. Several MSS techniques have been proposed in the literature for NARXs, including heuristic search, mathematical programming and sparsity promoting terms in overparametrized models (see e.g, 
\cite{CALAFIORE20143238} and references therein). Here, we focus on a two-step approach including $\ell 1$-norm relaxation of the $\ell 0$-norm followed by thresholding.

The paper is structured as follows: Section~2 details the developed SMNARX; Section~3 reports the developed algorithm for parameters estimation; Section~4 describes the adopted case study and summarizes the results achieved. 

%% file: problemFormulation.tex
\section{SMNARX identification problem}
Consider the class of hybrid dynamical systems defined in discrete time as follows:
\begin{align}
\begin{dcases}
y_k &= g_{1}(x_k; \theta_1)+ e_k, \ \ \text{when:} \ z_k=1\\
&...\\
y_k &= g_S(x_k; \theta_S)+ e_k, \  \text{when:} \ z_k=S\\
\end{dcases}
\end{align}
where $x_k=[y_{k-1},...,y_{k-n_a},u^T_{k-1},...,u^T_{k-n_b}]^T$ is a vector of past system outputs $y_k \in \mathbb{R}$ and exogenous inputs $u_k \in \mathbb{R}^q$ at discrete time $k\in \mathbb{Z}$, up to lags $n_a, n_b \in \mathbb{Z}^+$, and $e_k \in \mathbb{R}$ is an additive noise term. Each discrete mode $s \in {1,...,S}$ assumes a specific deterministic nonlinear mapping defined by a linear combination of basis functions $\left\lbrace \varphi_{i}(x_k)\right\rbrace_{i=1}^n $: 
\begin{equation}
g_s(x_k; \theta_s)=\sum_{i=1}^{n}\theta_{s,i} \varphi_{i}(x_k)
\end{equation}
As introduced above, in this work we adopt a polynomial NARX form in the states, hence the regressor vector $\varphi_{i}(.)$ is constituted by monomials of $x_k$ up to a fixed degree $n_d \in \mathbb{Z}^+$.
Considering a sequence of $N$ observed outputs $\mathcal{Y}_N=\left\lbrace y_k\right\rbrace_{k=1}^N$ and related lagged vectors $\mathcal{X}_N=\left\lbrace x_k\right\rbrace_{k=1}^N$, and following the first-order Markovian assumption over the latent modes:
\begin{equation}
p(z_k|z_{k-1},z_{k-2},...,z_{1},\mathcal{Y}_{N-1},\mathcal{X}_{N-1})=p(z_k|z_{k-1})
\end{equation}
the joint probability over the path of the hidden states $\mathcal{Z}_N=\left\lbrace z_k\right\rbrace_{k=1}^N$  factorizes as follows: 
\begin{equation}
p(\mathcal{Z}_N)=p(z_N,...,z_{k},...,z_{1})=\prod_{k=2}^{N}p(z_k|z_{k-1})p(z_1)
\end{equation}
while the joint mode/observation probability at $k$ results:
\begin{equation}
p(y_k,z_k|\mathcal{Z}_{k-1},\mathcal{X}_{k})=p(y_k|z_k,x_k)p(z_k|z_{k-1})
\end{equation}
The conditional likelihood of the observation sequence $\mathcal{Y}_N$ is obtained through marginalization over the hidden-state path:
\begin{equation}
\small p_\Theta(\mathcal{Y}_N|\mathcal{X}_N)=\sum_{\mathcal{Z}_N}\prod_{k=2}^N p_\theta(y_k|x_k,z_k)p_\mathcal{A}(z_k|z_{k-1})p_\theta(y_1|x_1,z_1)\Pi
\end{equation}
where $p_\mathcal{A}(z_k|z_{k-1})$ represents components of the transition probability matrix $\mathcal{A}=\left\lbrace a^{i,j}:=p(z_k=j|z_{k-1}=i)\right\rbrace_{i,j=1}^S $, $\Pi=\left\lbrace \pi^i\right\rbrace_{i=1}^S$, $\sum_{i=1}^{S}\pi^i=1$ the initial state probabilities and $p_\theta(y_k|x_k,z_k)$ the $\theta$-parametrized emission probabilities defined hereafter. $\Theta=\left\lbrace \theta,\mathcal{A},\Pi\right\rbrace$ summarizes the set of SMNARX model parameters to be estimated. 
The likelihoods of the output observations $y_k$ are assumed independent across the time instants given the state and the conditioning inputs at $k$. The shape of the emission distributions in the discrete states depends on the considered noise form. In this work, we follow the common Gaussian assumption $y_k|z_k=s,x_k \ \sim N(.; g_s(x_k; \theta_s), \sigma^2)$, obtaining a set of $S$ subsystems densities:
\begin{equation}
p(y_k|x_k,z_k=s, \theta_s, \sigma)=  \frac{1}{\sqrt{2\pi}\sigma}e^{-\frac{1}{2\sigma^2} \left(y_k - g_s(x_k; \theta_s) \right)^2}
\end{equation}   
where the NARX predictors provide the conditional means and $\theta=\{\left\lbrace \theta_s\}_{s=1}^S , \sigma \right\rbrace $ summarizes the submodels parameters. 

It is worth noting that, beside being a class of models for hybrid systems, SMNARX are a particular form of conditional probability Mixtures where component responsibilities depend on the assignments in the preceding time instant, referred to as Hidden Markov Experts in the time series analysis field \cite{RePEc:wsi:wschap:9789812810663_0002}. Besides, they constitute an extension of the Hidden Markov Models, widely exploited in the speech recognition context, including conditioned output distribution \cite{IOHMM}. If the Markovian conditioning over the mode switching mechanism is removed - hence loosing the strength of SMNARX reported above - the active state is estimated using the normalized emission probabilities. In fact, such simplification leads to conventional Mixture of Experts previously employed for PWARX and timed switching ARX systems identification \cite{JIN2012436}. 

The overall SMNARX identification given a collection of input/output data $\mathcal{D}=\left\lbrace (x_k, y_k)\right\rbrace_{k=1}^N$  is a highly difficult task involving discrete mode inference over the observed trajectories, the estimation of the transition probabilities and NARX parameters $\left\lbrace \theta_s\right\rbrace_{s=1}^S$, the selection of the components in the regressor vectors $\left\lbrace \varphi_i(.)\right\rbrace_{s=1}^S$, as well as model orders $n_a, n_b$ and the number of modes $S$. 
In this work, we follow the common assumption of predetermined number of modes and maximum order/lag, often inferred in practice by preliminary data analysis, cross validation and order selection techniques \cite{lauer_book}. 
Despite such simplification, the identification problem is still challenging, mainly due to the unknown active states and proper regressors shapes. To tackle this issues, we implemented the estimation method reported in the next section.

%% file: Method.tex
\section{Parameters estimation method}
\subsection{Expectation Maximization approach}
While a brute force maximization of the conditional likelihood $p_\Theta(\mathcal{Y}_N|\mathcal{X}_N)$ would be impracticable for realistic problem sizes, a much more efficient solution can be achieved under the Expectation Maximization (EM) framework. Constituting a general purpose technique for parameters estimation in case of missing data, EM can be employed to address the latent modes of switching Markovian models. Specifically, the likelihood optimization problem is approached by iterating over the following Expectation and Maximization steps:
\begin{align}
\begin{aligned}
E&-step: \\ \nonumber
&\mathcal{Q}\left( \Theta, \Theta^{(t)} \right)=\mathbb{E}_{\mathcal{Z_N}} \left[ p(\mathcal{Y}_N,\mathcal{Z}_N|\mathcal{X}_N, \Theta) |\mathcal{Y}_N,\mathcal{X}_N, \Theta^{(t)} \right] \\
M&-step: 
\end{aligned}
\end{align}
\begin{equation}
\Theta^{(t+1)}=\underset{\Theta}{\small{argmax}} \ \mathcal{Q}\left( \Theta, \Theta^{(t)} \right)
\end{equation}
where $\mathcal{Q}(\Theta, \hat{\Theta})$ is a deterministic auxiliary function representing the expectation over the latent states of the complete data likelihood (i.e, assuming the latter to be given) determined using the model parameters from the previous iteration $t$. Hence, the $\mathcal{Q}$-function evaluation involves the computation of the posterior distribution of the latent states, as shown here: 
\begin{equation}
 \mathcal{Q}(\Theta, \Theta^{(t)})=\sum_{\mathcal{Z_N}}p(\mathcal{Z}_N|\mathcal{Y}_N,\mathcal{X}_N, \Theta^{(t)}) p(\mathcal{Y}_N,\mathcal{Z}_N|\mathcal{X}_N, \Theta) \nonumber
\end{equation}
Practically, it works as a kind of "approximated oracle": soft-labels are assigned to latent states over the trajectories by averaging over the posterior, updated using the parametrization obtained from the previous maximization phase, starting from initial guesses\cite{IOHMM}. 
The algorithm iterates until convergence, tested by stopping conditions on the likelihood or maximum number of iterations.
As common for EM, several initialization are required, by sampling different values of the parameters to tackle eventual convergence to poor local minima. Notably, EM has interesting convergence properties, holding also when a closed form solution is not available for the M-step \cite{Bishop:2006:PRM:1162264}. In this case, an increase of the $\mathcal{Q}$-function is required in each iteration, leading to the Generalized EM algorithm (GEM). 

\subsection{Computation of the latent state posterior in the E-step}

As the probability graph of the SMNARX has a tree structure similar to Hidden Markov Models (and Hidden Markov Experts), the posterior can be computed efficiently through a slight modification of a widely adopted message passing algorithm, often referred to as forward-backward or Baum-Welch algorithm \cite{Bishop:2006:PRM:1162264},\cite{RePEc:wsi:wschap:9789812810663_0002}. 
Specifically, two auxiliary variables are introduced to recursively perform the estimation step, namely the forward and backward variables.
By introducing a specific term for the probability of being in state $s$ at $k$ after seeing the first t observations (given the model parameters): 
\begin{equation}
p(z_k=i, y_k,...,y_1):= \alpha_k^i
\end{equation}
the probability of the overall sequences of observations is obtained by marginalizing the states occurring at the final step:
\begin{equation}
p(\mathcal{Y}_N)=\sum_{i=1}^{S}p(z_N=i, y_N,..,y_k,..,y_1)=\sum_{i=1}^{S}\alpha_N^i
\end{equation}
We hided explicit references to the conditioning parameters to lighten notation.
$\alpha_k^i$, typically referred to as the forward variable in the HMM context, can be computed efficiently using a recursive procedure, starting from initial assignments $\alpha_1^i=\pi^ib_1^i$, through the following factorization:
\begin{align}
\begin{aligned}
\alpha_{k}^j= &\sum_{i=1}^{S} [p(z_{k-1}=i, y_{k-1},...,y_1)p(z_k=j|z_{k-1}=i)\\
&* p(y_{k}|x_{k},z_k=j)]=
\sum_{i=1}^{S}\left( \alpha_{k-1}^i a^{i,j}\right) b_k^j
\end{aligned}
\end{align}
where $b_k^j=p(y_{k}|x_{k},z_k=j)$ represents the emission probability given the active state $z_k=j$. 
It is worth noting that such operations result linear in time, as opposed to the exponential increase with the sequence length given by full paths coverage. 
The sequences are processed in the opposite direction by the backward variables, representing the joint probability of the observations from $k+1$, given the active state at $k$:
\begin{equation}
p(y_{N},...,y_{k+1}|z_k=i):= \beta_k^i
\end{equation}
computed through the following recursion, starting from $N$:
\begin{align}
\begin{aligned}
\small \beta_{k}^i= &\sum_{j=1}^{S} [p(z_{k+1}=j|z_{k}=i)p(y_{k+1}|x_{k+1},z_{k+1}=j)\\
&*p(y_{N},...,y_{k+2}|z_{k+1}=j)]=
\sum_{j=1}^{S}a^{i,j}b_{k+1}^j \beta_{k+1}^j
\end{aligned}
\end{align}
Hence, the overall observation path (summarized by $\alpha_k^s,\beta_k^s$) is employed to compute the state posterior at time $k$: 
\begin{equation}
p(z_k=j|\mathcal{Y}_N)=\frac{\alpha_k^j \beta_k^j}{\sum_{s=1}^{S}\alpha_k^s \beta_k^s}:= \gamma_k^j
\end{equation}
while the joint posterior distribution of two successive latent states over the path is obtained by:
\begin{equation}
p(z_k=i, z_{k+1}=j|\mathcal{Y}_N)=\frac{\alpha_k^i  a^{i,j} b_{k+1}^j \beta_{k+1}^j}{\sum_{s=1}^{S}\alpha_k^s \beta_k^s}:= \xi_k^{i,j}
\end{equation}

As the computations of the auxiliary variables for recursive estimation include also future observations over the training sequence (by $\beta_{k+1}^j$), the state posterior $\gamma_k^j$ cannot be considered for prediction in testing conditions. To this end, the following predictive mode probabilities can be adopted:
\begin{equation}
p(z_{k}=j|\mathcal{Y}_{k-1})=\frac{\sum_{i=1}^{S}\alpha_{k-1}^i a^{i,j}}{\sum_{i=1}^{S}\sum_{j=1}^{S}\alpha_{k-1}^i a^{i,j}}:=f_{k}^j
\end{equation}
inferring the latent state at $k$ by iterating the forward variable till $k-1$ (giving the modes probabilities in the previous instant), times the consequent transitions. 
The predicted means are then obtained by combining the modes through $f_k^s$: 
\begin{equation}
\hat{y}_k=\sum_{s=1}^{S}f_k^s\sum_{i=1}^{n}\theta_{s,i} \varphi_{i}(x_k)
\end{equation}

On the other hand, the posteriors $\gamma_k^s, \xi_k^{i,j}$ provided by the forward-backward procedure are employed to update the parameters in the M-step, as detailed in the next subsection. 

\subsection{Model parameters updates in the M-step}

The probability matrix is computed by
the expected number of transitions from state i to state j on the overall the expected number of transitions from state i. 
\begin{equation}
\hat{a}^{i,j}=\frac{\sum_{n=1}^{N-1}\xi_k^{i,j}}{\sum_{k=1}^{N-1}\sum_{s=1}^{S}\xi_k^{i,s}}
\end{equation}
while the initial probability parameter is assigned to $\pi^i=\gamma_1^i, \ \forall i$.
The variance parameter is computed by:
\begin{equation}
\hat{\sigma^{2}} = \sum_{s=1}^{S}\sum_{k=1}^{N} \gamma_k^s \left(y_k - \sum_{i=1}^{n}\theta_{s,i} \varphi_{i}(x_k) \right)^2/\sum_{s=1}^{S} \sum_{k=1}^{N}\gamma_k^s
\end{equation}
The update of the NARXs parameters result in the following set of subproblems, each weighted by the related sample-wise soft-assighments to the modes $\gamma_k^s$ during the iterations:
\begin{equation}
\mathcal{L}_s(\theta_s)=\sum_{k=1}^{N}\gamma_k^s (y_k - \sum_{i=1}^{n}\theta_{s,i} \varphi_{i}(x_k))^2, \ \forall s \in S
\end{equation}
Notably, discrete mode classification and NARX regression tasks are disentangled within the EM iterations. 

As reported in Section I, automated mechanisms has to be introduced to select the most suitable NARX components (in terms of contribution to the prediction performances), addressing the curse of dimensionality of polynomial expansions as well as the increased overfitting potential of overparametrized models. Hence, the $\theta_s, \forall s \in S$ are assumed to attain specific sparse patterns (i.e, most coefficients equal zero) as the corresponding basis terms do not provide sensible contributions to the predictions, resulting in parsimonious models.

To this end, the best subset estimation approach considers a $l0$-norm on the parameters vector, thus constraining the regression fitting with at most $\kappa$ non-vanishing components.
Considering the specific characteristics of the SMNARX model, this leads to a set of $S$ sub-problems of the form: 
\begin{equation}
\underset{\theta_s}{\small{min}} \ \mathcal{L}_s(\theta_s) \ s.t. \ ||\theta_s||_0 \leq \kappa; \ ||\theta_s||_0 = \sum_{i=1}^{n}\mathbf{1}\left\lbrace \theta_{s,i}\neq 0\right\rbrace 
\end{equation}
A sensible amount of studies has been dedicated to the achievement of scalable and computational efficient solution of such NP-hard problems \cite{Yin2020}. Popular techniques include  
greedy forward/backward stepwise selection \cite{hastie2017extended}, randomized algorithms \cite{BIANCHI2020100818}, Mixed Integer Programming based methods \cite{Bertsimas} and convex relaxations through bridge regression \cite{wang2017bridge}. 

In this work we focus on the latter class, leaving the investigation of alternative approaches to future developments.
Specifically, we consider the closest convex surrogate of the $l0$-norm, i.e, the $l1$-norm (often referred to as least absolute shrinkage and selection operator - LASSO), widely employed in the machine learning field to introduce sparsity priors and regularization in high dimensional settings \cite{Goodfellow-et-al-2016}.
Adopting the conventional Lagrangian form, the discrete mode weighted regressions are augmented by sparsity promoting terms: 
\begin{equation}
\underset{\theta_s}{\small{min}} \ \sum_{k=1}^{N}\gamma_k^s \left(y_k - \sum_{i=1}^{n}\theta_{s,i} \varphi_{i}(x_k) \right)^2 + \lambda \sum_{i=1}^{n}|\theta_{s,i}|
\end{equation}
where $\lambda \geq 0$ is a tuning hyperparameter weighting the regularization.
Practically, such penalized regression form encourages sparse solutions while performing parameters estimation, resulting in a soft-thresholding behavior, as opposed to the hard-thresholds of the $\ell0$-norm based subset estimation. 
While LASSO can theoretically uncover sparse structures 
under specific strong conditions, it suffers several limitations in finite sample settings due to its intrinsic downward bias caused by the shrinkage action \cite{Yin2020}. 
Hence, several variants have been investigated to obtain weaker requirements than the irrepresentable condition of the conventional LASSO for variable selection consistency \cite{wang2017bridge}.
In this work, we focus on the two stage selection scheme developed in \cite{Meinshausen}, post-processing the parameters vectors computed by the $\ell1$-regularized regressions component-wise by the hard-thresholding function (i.e., $\ell1$2S):
\begin{equation}
\small \bar{\theta}_s=\eta_0(\hat{\theta}_s; \upsilon^2/2), \text{with: } \eta_0(x;\varUpsilon )=x\mathbf{1}\left\lbrace|x|\geq \sqrt{2\varUpsilon} \right\rbrace  
\end{equation}
where $\hat{\theta}_s$ represent the parameters estimate obtained during each EM iteration by the solution of (21) and $\upsilon \geq0$ is a threshold hyperparameter. Hence, sign consistency can be achieved under weaker requirements, while the first stage is exploited to provide good approximations in the $\ell2$-sense. A detailed comparison of two-stage variable selection techniques to LASSO under various Signal to Noise ratios is reported in \cite{wang2017bridge}, showing that the optimal configuration of the regularization weight in the bridge regression step is the same for estimation and selection. The sparse patterns are then recovered under the assumption that proper nonzero coefficients are sufficiently large (in absolute value), controlling the discovery trade-off by tuning $\upsilon$  \cite{Meinshausen}. Within the different approaches that can be exploited for hyperparameters calibration (e.g, combining the two-step procedure with the knockoff framework), in this work we adopt the decreasing search procedure deployed in \cite{wang2017bridge} with configurable window/grid size through cross-validation. 
Considering the specific characteristics of the SMNARX identification problem at hand, requiring both parameter estimation, selection and discrete mode inference, we found useful experimentally (as shown in the next section) to perform a burn-in period using only the $\ell1$-norm to get rid of the soft-assignments performed during the first parts of the EM iterations. Hence, the hard-thresholding step is introduced once the likelihood reach stable conditions, controlled by a parametrized tolerance. The $\ell1$ action is then modulated by the selected features.
To solve the 
weighted regressions, we adopted cyclical coordinate-wise minimization \cite{Tibshirani}.


%% file: Results.tex
\section{Numerical results}

To the best of our knowledge, this work constitute the first investigation of Switched Markov Polynomial NARX systems in the literature, and a benchmark problem for this  systems has not been proposed yet. Hence, we constructed the following system by merging the characteristics of the SNARX problem deployed in \cite{BIANCHI2020100818} (aimed to investigate switched polynomial NARX) with the SMARX system considered in \cite{JIN2012436}.
The resulting SMNARX system to be identified is composed by three different nonlinear sub-models with specific regressors:

\begin{align}
\begin{dcases}
\nonumber
y_k&=0.5y_{k-1} + 0.8u_{k-2} +u_{k-1}^2 -0.3y_{k-2}^2 + e_k\\
y_k&=0.2y_{k-1}^3 - 0.5y_{k-2} - 0.7y_{k-2}u_{k-2}^2 + 0.6u_{k-2}^2 + e_k\\
y_k&=0.5 y_{k-2} -0.4 y_{k-1} + 0.2 u_{k-1} -0.4 u_{k-3}y_{k-1} + e_k
\end{dcases}
\end{align}
with $e_k \sim N(0,0.1^2)$ and input sampled from a uniform distribution $u_k \sim U[-1,1]$. The transition probability matrix governing the Markovian discrete mode switching has been defined as $A=[[0.98,0.02,0],[0,0.98,0.02],[0.02,0,0.98]]$.

We generated a sequence of 12000 samples, thus including a sensible number ($>$ 300) of state switches in the identification set. The first 10000 has been devoted to training, the successive 1000 to validation and the last 1000 constitute the test set.
Following \cite{BIANCHI2020100818}, we set the orders to oversized values $n_a,n_b=4,n_d=3$ resulting in 165 regressors.
Then, we structured the training sequence into mini-batches of length 200.

Both the overall model architecture and the EM-based estimation algorithm have been developed in Numpy-1.18. The weighted least squares subproblems, including the $\ell1$-norm and thresholding steps are implemented using the LinearModel and SelectFromModel classes of Scikit-learn-0.23. 

To avoid convergence to poor local minimum (where the states collapse to common, i.e., average, modes), we found useful experimentally to initialize $ \gamma_k^s \sim U[0.31,0.35], \forall s,k$, while constraining a valid distribution parametrization. Moreover, the EM is executed (for each hyperparameters configuration) by starting from 10 random initialization and then selecting the run with the higher likelihood. The tolerances on the likelihood improvements have been set to $1e^{-2}$ and $1e^{-6}$ to control burn-in and training stops respectively, including a maximum number of iterations of 100. Prediction accuracy is measured by the conventional Root Mean Squared Error (RMSE). 
Using the decreasing grid search procedure in cross-validation over a window $[1e^{-6}, 1e^{1}]$, $\lambda=5e^{-4}$ resulted a reasonable choice for test purpose. 
Table \ref{tab:tab1} reports the obtained validation RMSE for the most significant subset of $\lambda$ in the search window. The threshold have been set to $\upsilon=5e^{-2}$.

To evaluate the accuracy of the SMNARX  reconstruction we employed the performance measures adopted in \cite{PILLONETTO201621}\cite{JIN2012436}: 
\begin{equation}
\mathcal{F}_{\theta}= \frac{1}{S}\sum_{s=1}^{S} \left(1-\frac{\lVert \theta_s - \hat{\theta}_s \lVert}{\lVert \theta_s \lVert} \right), \ \ \mathcal{F}_{A}=1-\frac{\lVert \hat{A} - A \lVert}{\lVert A \lVert}
\end{equation}
namely the parameters and transition probability indexes.  To investigate the mode estimation performance, we adopted the index
$$ \mathcal{F}_{s}=1/N_{t}\sum_{k=1}^{N_{t}} \mathbf{1}_{\hat{s}_k=s_k}$$ 
where $\hat{s}_k$ and $s_k$ represents the inferred and true state at $k$ while $N_t$ the train/test size, where the subscript $s_{tr} (s_{te})$ in Table \ref{tab:tab2} refers to training (test) set.\\
As the order of the discrete states in the model do not necessarily match that of the target system, post-processing is required before evaluation \cite{handbook_mixtures}. As suggested in \cite{PILLONETTO201621}, without loss of generality, we reordered the sub-models estimate by the Euclidean norm to the target parameters. 
Following \cite{HARTMANN20159}, statistics are computed over 100 runs of the estimation algorithm on randomly sampled sequences using the same hyperparameters.
Under the adopted configuration, convergence have been observed in approximately 20 iterations on average. The mean computation time of one iteration on a CPU-i7-2.5-GHz-RAM-8Gb for this case study has been 0.8 seconds.
\begin{table}[t]
	\caption{Validation RMSE over different $\lambda$}
	\begin{center}
		\begin{tabular}{|c|c|c|c|c|c|c|c|}
			\hline			 
			$\lambda$
			& 1e-1
			& 5e-2
			& 1e-2
			& 5e-3
			& 1e-3
			& 5e-4
			& 1e-4\\
			\hline		
			& 0.5013
			& 0.3382
			& 0.2007
			& 0.1757
			& 0.1692
			& 0.1691
			& 0.1691\\
			\hline	
		\end{tabular}
		\label{tab:tab1}
	\end{center}
	\caption{Performance indexes and parameters statistics (100 runs)}
	\begin{center}
		\begin{tabular}{|c|c|c|c|c|c|}
			\hline
			\textbf{\textit{ }}
			& \textbf{\textit{$\mathcal{F}_{s_{tr}}$}}
			& \textbf{\textit{$\mathcal{F}_{s_{te}}$}} 
			& \textbf{\textit{$\mathcal{F}_{A}$}}
			& \textbf{\textit{$\mathcal{F}_{\theta}$}} 
			& \textbf{\textit{$n_{feat}$}}
			\\
			\hline			 
			SMNARX-EM
			& 0.995
			& 0.966
			& 0.995
			& 0.941
			& 165\\
			\hline
			SMNARX-EM-$\ell1$
			& 0.996
			& 0.967
			& 0.995
			& 0.969
			& 165\\
			\hline	
			SMNARX-EM-$\ell1$2S
			& 0.996
			& 0.967
			& 0.995
			& \textbf{0.990}
			& \textbf{4}\\
			\hline	
		\end{tabular}
		\label{tab:tab2}
	\end{center}
	\begin{center}
		\begin{tabular}{|c|c|c|c|c|c|}
			\hline
			$\ell1$2S
			& \textbf{\textit{$Mean$}}
			& \textbf{\textit{$Std$}} \\
			\hline			 
			$\theta_1$
			& [0.500, 0.800, 0.999, -0.299]
			& [2.3e-3, 3.1e-3, 4.2e-3, 1.9e-3]\\
			\hline
			$\theta_2$
			& [0.200, -0.500, -0.698, 0.599]
			& [5.3e-3, 6.0e-3, 1.4e-2, 4.0e-3]\\
			\hline
			$\theta_3$
			& [0.498, -0.400, 0.200, -0.399]
			& [6.7e-3, 7.7e-3, 2.9e-3, 9.3e-3]\\
			\hline
			$\sigma^2$
			& 0.010
			& 1.3e-4\\
			\hline	
		\end{tabular}
		\label{tab:tab3}
	\end{center}
\end{table}
\begin{figure}[t]
	\includegraphics[width=1\linewidth]{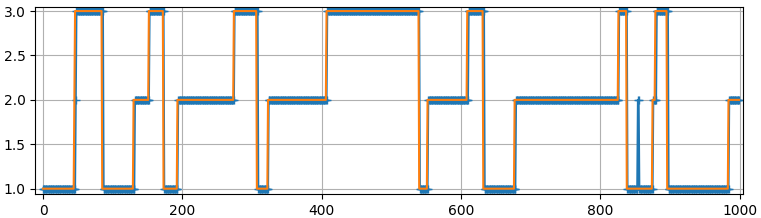}
	\caption{Predicted (blue) vs True (orange) modes over the test set}
	\label{fig:w_hist}
\end{figure} 

Table \ref{tab:tab2} reports the measured indexes and the submodels parameters statistics computed over 100 runs, comparing the integration of the shrinkage action and the thresholding step. 
Notably, the EM-based estimation technique achieved a proper classification of the active states and switching patterns, as depicted also by Figure 1, reporting the predicted modes over the test set.   
Furthermore, the integration of the two stage selection utility enabled the extraction of the correct set of features ($n_{feat}$) - for each submodel - from the sparsified patterns provided by the $\ell1$-norm bridge estimator, and the consequent improvement of NARXs submodel parameters fitting ($\mathcal{F}_{\theta}$). To provide further insights into such behavior, we report in Figure 2 the plots of the coefficients paths of the NARX over the iteration before/after the threshold insertion on the conventional LASSO, starting from random initialization. 
Visibly, the base $\ell1$-norm sparsifier, employed during the burn-in phase, provides foreseeable approximations \cite{Meinshausen}, where false positive terms have been assigned with a very small parameter value. Once the threshold phase is activated, such minor components are incrementally removed, while the coefficients of the remaining parameters are slightly adjusted accordingly, until convergence. 

Clearly, a proper hyperparameter configuration is crucial to achieve such behavior. While the adopted cross-validation based procedure provided suitable results on this case study, this topic would require further investigation. Chiefly, enhanced heuristics/search techniques might further speed-up tuning, fostering scalability to increasing problem complexity.
   
\begin{figure}[t]
	\includegraphics[width=0.95\linewidth]{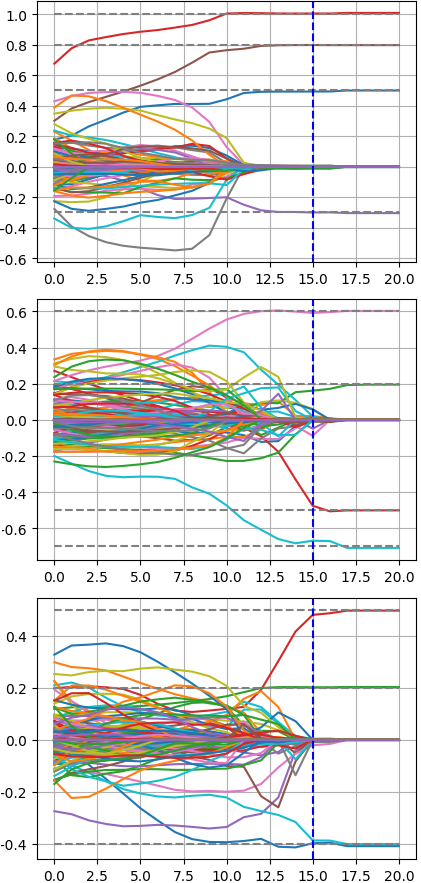}
	\caption{NARXs parameters during iterations for each mode}
	\label{fig:w_hist}
\end{figure}

%% file: Conclusion.tex
\section{Conclusion and Next Steps}
\label{Conclusion}
In this work we investigated the identification of Switched Markov Nonlinear ARX system with  parameterized polynomial expansions. To tackle the challenging brute force likelihood maximization, estimation is performed using an Expectation Maximization algorithm, disentangling discrete mode classification and NARXs regression tasks. Sparse features patters are enforced through a two stage approach including a $\ell1$-norm based bridge estimation. Numerical analysis has been performed on a benchmark SMNARX problem, showing the capability to achieved a proper identification of the active states, switching patterns, and submodel parameters. 

Next developments will include the integration of enhanced search techniques for hyperparameters tuning, the investigation of further model structure selection approaches, modes/order selection methods, and the application to a real case study.
%